\documentclass[conference]{IEEEtran}
\IEEEoverridecommandlockouts
\usepackage{cite}
\usepackage{amsmath,amssymb,amsfonts}
\usepackage{algorithmic}
\usepackage{graphicx}
\usepackage{textcomp}
\usepackage{xcolor}
\usepackage{hyperref}
\def\BibTeX{{\rm B\kern-.05em{\sc i\kern-.025em b}\kern-.08em
    T\kern-.1667em\lower.7ex\hbox{E}\kern-.125emX}}
\begin{document}

\title{Adaptive Growth: Real-time CNN Layer Expansion 

}

\author{\IEEEauthorblockN{%1\textsuperscript{st}
Yunjie  Zhu}
\IEEEauthorblockA{\textit{School of Computing} \\
\textit{University of Leeds}\\
sc20yz@leeds.ac.uk}
\and
\IEEEauthorblockN{%2\textsuperscript{nd} 
Yunhao Chen  }
\IEEEauthorblockA{\textit{School of Artificial Intelligence and Computer Science} \\
\textit{Jiangnan University}\\
1191200221@stu.jiangnan.edu.cn }
\footnotesize \textsuperscript{*}Note: equally contributed
} 

\maketitle

\begin{abstract}
Deep Neural Networks (DNNs) have shown unparalleled achievements in numerous applications, reflecting their proficiency in managing vast data sets. Yet, their static structure limits their adaptability in ever-changing environments. This research presents a new algorithm that allows the convolutional layer of a Convolutional Neural Network (CNN) to dynamically evolve based on data input, while still being seamlessly integrated into existing DNNs. Instead of a rigid architecture, our approach iteratively introduces kernels to the convolutional layer, gauging its real-time response to varying data. This process is refined by evaluating the layer's capacity to discern image features, guiding its growth. Remarkably, our unsupervised method has outstripped its supervised counterparts across diverse datasets like MNIST, Fashion-MNIST, CIFAR-10, and CIFAR-100. It also showcases enhanced adaptability in transfer learning scenarios. By introducing a data-driven model scalability strategy, we are filling a  void in deep learning, leading to more flexible and efficient DNNs suited for dynamic settings. Code:(https://github.com/YunjieZhu/Extensible-Convolutional-Layer-git-version).

\end{abstract}


\section{Introduction}
Inspired by the human visual cortex, Convolutional Neural Networks (CNN) are proposed and proven to be a powerful model for various computer vision tasks, such as image classification, object detection, and image segmentation. \cite{b1,b1_2,b1_3,b1_4,b1_5} The success of such models, along with the development of a richer dataset and network architectures, has brought various advanced image-based applications, ranging from self-driving cars to medical image analysis and beyond. \cite{b2,b2_11,b2_12,b2_13,b2_14,b2_15,b3,b4}

Despite their widespread success, CNNs are not without limitations. \cite{limitation1,limitation2,limitation3,limitation4,limitation5} One of the significant constraints is the predefined size of the Convolutional layer, which severely restricts the network's adaptability to dynamic environments. This rigidity hampers the network's ability to adjust to shifts in data distribution or to acquire new skills without undergoing a complete retraining process—a procedure that is both computationally intensive and time-consuming. Hence, various methods are proposed, including transfer learning, incremental learning, and meta-learning, and have shown great success under various scenarios. \cite{b2_1, b2_2, b2_3, b2_4, b2_5, b2_6} 

Dynamically expanding a layer within a Deep Neural Network (DNN) can offer a more flexible and adaptive learning system. Such a growth mechanism would naturally augment the network's capacity, facilitating easier task allocation by creating new bits of output from the layer. For example, previous works have shown that adding new experts for new tasks could improve overall performance and avoid forgetting problems. \cite{b3_1, b3_2, b3_3, b3_4}  However, most of these methods require retraining an entire network and supervision, leading to a passive adaption to environments. \cite{b2_n_1, b2_n_2}

In this study, we explored the proactive, dynamic expansion of a convolutional layer, adapting based on real-time image input by gauging its responsiveness. We evaluated layers crafted using our novel algorithm against conventionally supervised-trained layers across four datasets: MNIST, Fashion-MNIST, CIFAR10, and CIFAR100.\cite{dataset1, dataset2, dataset3, dataset4} The upcoming sections will delve into our algorithm's methodology, present our experimental findings, and contemplate potential avenues for future research.

\section{Problem Statement and Related Works}
The problem in our work is how to enable the unsupervised expansion of a convolutional layer with changing environments. 
Our algorithm shares a akin design principle with Adaptive Resonance Theory (ART), which involves incorporating new neurons into the network as fresh data arrives.\cite{b3_2, b3_2_2} However, our proposed algorithm generates a convolutional layer that is compatible with modern DNNs as a module, which differs from the ART-based generation models that are independent classification models.

\section{Method}

\subsection{Procedure}
\begin{figure*}[htbp]
\centerline{\includegraphics[scale=0.38]{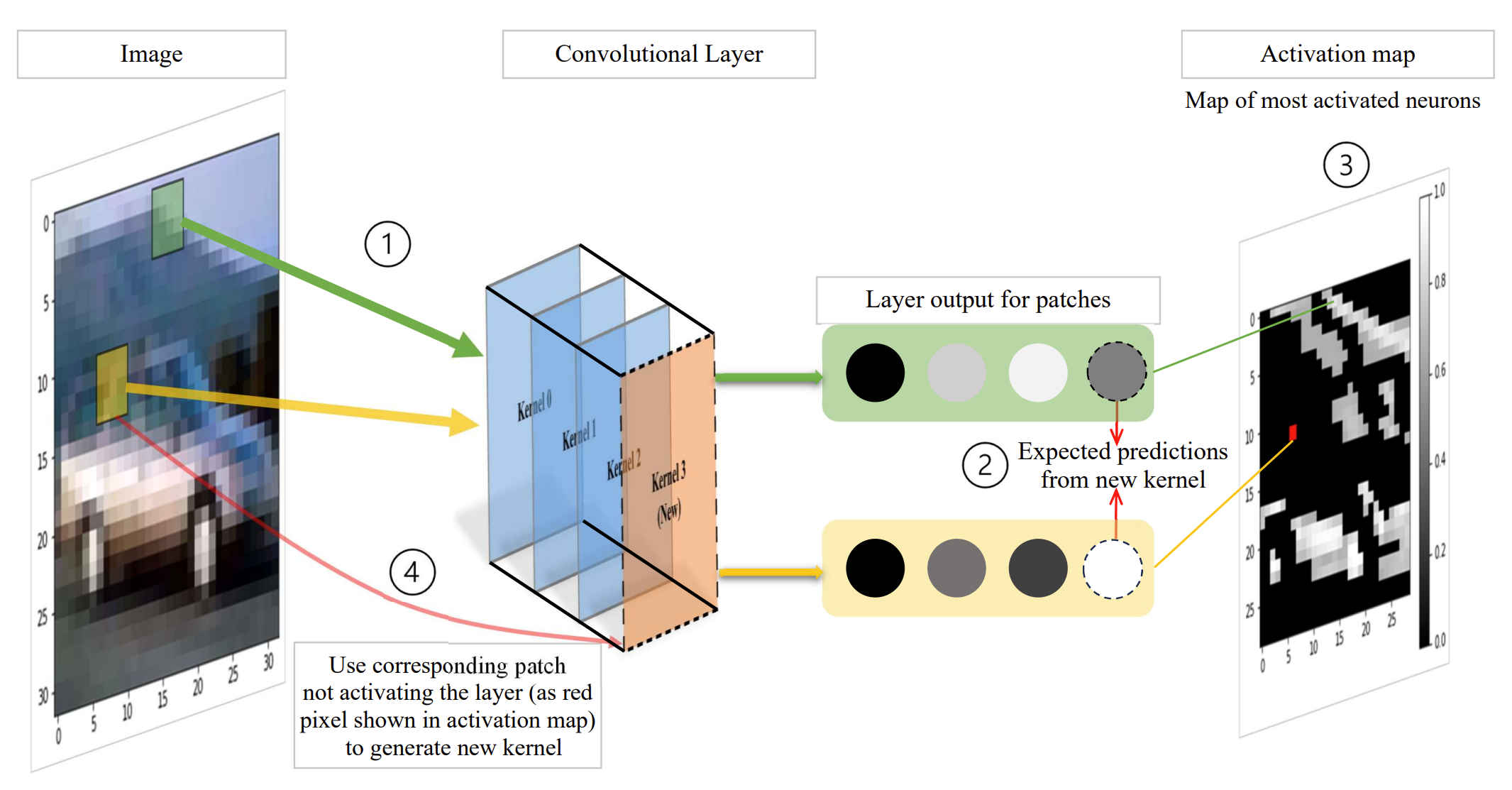}}
\caption{Procedure for generating a new kernel. \textcircled{1} Input the image into the designated layer. \textcircled{2} Extract the output vector from the kernels in the layer corresponding to the specific image patch. Each circle is a float value of a kernel's output, ranging from 0 to 1, corresponding to a colour space from black to white. \textcircled{3} Generate the activation map, documenting the layer's responses, which helps the algorithm to identify the patches of image rejected by the layer. \textcircled{4} For image patches that do not activate any existing kernel in the layer, create a new kernel. This new kernel is designed to recognize and accept the selected patch $P_S$, producing an output of 1, while rejecting other patches with an output of 0. }
\label{Procedure_graph}
\end{figure*}
Our approach allows a convolutional layer to proactively grow based on data by two major processes: Measuring the layer’s responsiveness to images and kernel generation. When generating the kernel, our algorithm first processes the provided image’s patch to initialize a weight for the kernel, and then the weights are calibrated by training based on previously generated kernels. The procedure is demonstrated with the Fig.~\ref{Procedure_graph}

In Fig.~\ref{Procedure_graph}, four steps are taken to generate a new kernel based on the responsiveness of the layer. We define a kernel that accepts inputs by giving outputs above a defined threshold and rejects the input by giving a lower output. Hence, the responsiveness of the layer is measured by recording the highest values from the outputs of kernels for each patch of the image above the threshold, which yields the activation map. 

\subsection{Metric to Measure Layer’s Responsiveness – Inactive ratio}
In order to measure the responsiveness of a convolutional layer to decide whether the layer should be expanded, we calculate how many portions of the image activate the layer with the following equations for inactive ratio $H(X)$:

\begin{equation}
A(X_{i,j}) = max(C(X_{i,j})>\alpha) \label{eq_Act}
\end{equation}

\begin{equation}
H(X) = \frac{{M \times N - \sum \limits_{i=1}^{N} \sum \limits_{j=1}^{M} A(X_{i,j})}}{{M \times N}}
\end{equation}

Where $\alpha$ is a predefined threshold indicating the activation of a kernel in the layer, $M$ and $N$ are the width and height of a given image, and $C$ is the convolutional layer that accepts an image or a patch of the image as an input. The activation map $A(X_{i,j})$ measures if the patch of the image at coordinates $i, j$ activates an arbitrary kernel by giving an output above a predefined threshold $\alpha \approx \sigma(0.5)$, in which $\sigma$ is the activation function sigmoid. The activation map, which measures the responsiveness of a convolutional layer during generalisation (\textcircled{3} in Fig.~\ref{Procedure_graph}), is produced by feeding the entire image to function $A(X)$.

In actual implementation, the proposed algorithm will be stopped if $H(X)$ is lower than 0.1 for a batch of images used for generating the layer to avoid overfitting after most of the patterns have been learned.

\subsection{Initialising new layer}
To initialise the layer from void with the first set of images, we have designed the first kernel to recognize a patch of image $P_0$ with only 0 to recognize the fully black patches. The weights $W$ are set to uniformly -2, and bias $B$ is set to 1. Then, a set of kernels is initialised with the process in the following section with only loss on their assigned patch. The following Fig.~\ref{Act_map_init} demonstrates the responses of the initialised layer after this process.

\begin{figure}[htbp]
\centerline{\includegraphics[scale=0.35]{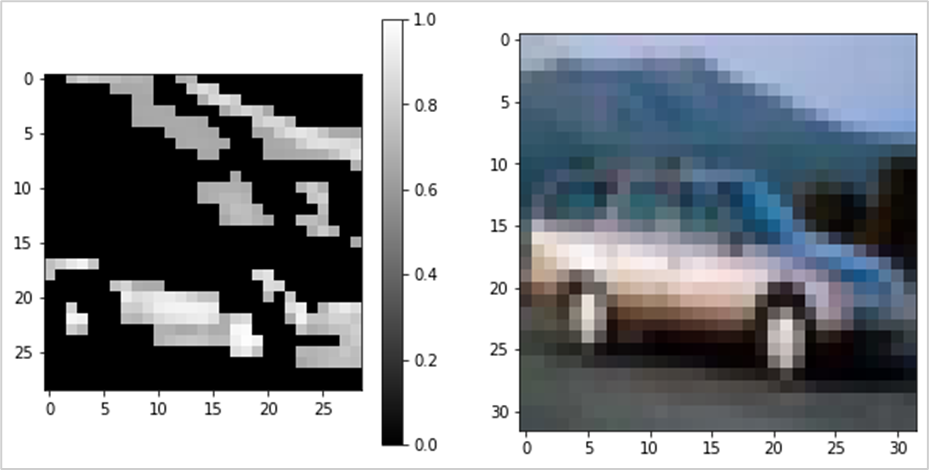}}
\caption{Activation map of initialised layer. A whiter pixel on the left represents higher activity above the threshold from the layer.}
\label{Act_map_init}
\end{figure}

\subsection{Initializing new kernel}\label{AA}

With a selected patch $P_S$ of the image not activating any kernel, we select it to generate the new kernel. We first centre the patch's values to 0, which avoids biases of outputs with:
\begin{equation}
\Omega =P- \overline{P}
\end{equation}

Then, a matrix $C$ could be obtained by scaling the $\Omega $. This operation sets the output of the kernel to 1 for the given patch without applying the activation function:
\begin{equation}
C=\frac{\Omega } {\sum_{i=1}^{N} \sum_{j=1}^{M} \left( \Omega  \circ P \right)_{i,j}}
\end{equation}

Where $\circ$ is the Hadamard product of two matrices.

However, $C$ might have extremely large values when the standard deviation is too small, which means it's a patch of image containing a more uniform colour, causing overfitting problems in training. We addressed this case by iteratively increasing the standard deviation of $\Omega $ with the following steps.
First, the positive and the negative values are extracted from the matrix into two matrices: $\Omega =\Omega ^++\Omega ^-$. Then, the following operation based on element-wise multiplication is applied to bring a higher standard deviation over $\Omega $:
\begin{equation}
\Omega ^{\prime}=\Omega ^{+} \circ \Omega ^{+}+\Omega ^{-} \circ \Omega ^{-}
\end{equation}

Finally, the weights and bias of a new kernel are:
$(W=C,bias=0)$ 
This weight and bias initialisation method, which extracts the information from the given patch of the image, will make the new kernel $K_{new}$ in the convolutional layer give:
$(K_{new} (P)=\sigma(1))$
Where $\sigma$ is the activation function selected for the convolutional layer.

For simplicity in design, the activation function implemented in the experiments is sigmoid. A potential adjustment needs to be made for other activation functions to measure whether a kernel responds to the given patch of the image by setting an upper limit in \eqref{eq_Act} when obtaining the activation map. The reason for our initialization method will be discussed in the subsequent section. 

\subsection{Kernel Generalization}

When training the new kernels after initialisation, the proposed algorithm will fetch the patches from the given image that have activated existing kernels with the highest value above the threshold $\alpha$. These patches are designated as negative samples and labelled 0 to be rejected by the new kernel. A problem arises when only one patch is labelled 1, and multiple negative samples are labelled 0 during training, which can result in the rejection of the $P_S$ for the new kernel. Hence, we have initialized the kernel's weights and biases using the $P_S$ in the previous section.

Then, the new kernel is generalized by training against the $P_S$ used for initialization and the negative samples $S^{-}$. The loss function $L$ implemented in our experiments is based on the Mean Squared Error (MSE) in \eqref{eq_Loss}. 
\begin{equation}
L=MSE(F_{new}(P)-\lambda)+MSE(F_{new}(S^{-})) \label{eq_Loss}
\end{equation}

After training, the new kernel will be generalized to accept $P_S$ containing the desired pattern by giving output as 1, while rejecting as many negative samples as possible by giving a 0. Considering that there might be no negative samples for the first few kernels, we set $\lambda = \alpha$ to avoid overfitting, which produces several example kernels at the beginning of generation.

\begin{figure}[htbp]
\centerline{\includegraphics[scale=0.32]{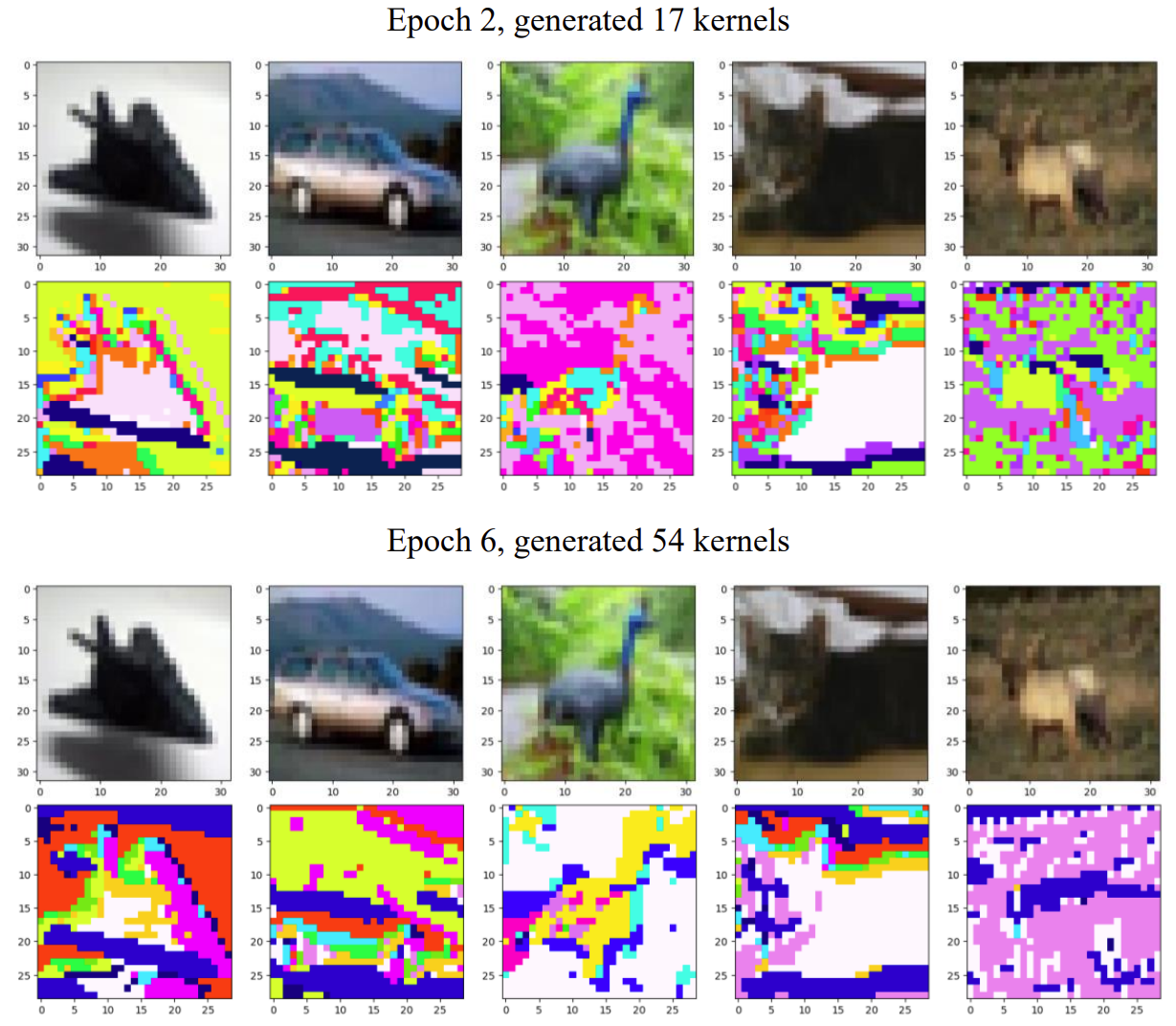}}
\caption{Example of layer activity at 2nd and 6th epochs.}
\label{Example_vis}
\end{figure}

With the above processes, the layer will gradually learn patterns in images. Fig.~\ref{Example_vis} shows the layer's responses to different images with 17 kernels in the 2nd epoch and 54 kernels in the 6th epoch.

In Fig.~\ref{Example_vis}, different colours represent different kernels having the highest activity at the pixel in the feature map. Some kernels' dominance might be replaced by others, which is acceptable due to the potential hierarchy relation between patterns in images.
\section{Experiment}
We have designed several scenarios to test the functionality and effectiveness of the convolutional layer generated by our algorithm. These experiments focus on the ability to expand the layer's size with new incoming features and the performance of the generated layer in classification tasks.

To compare the counterparts of generated convolutional layers, we have implemented lightweight models Model 1 and Model 2 with architectures in TABLE~\ref{tab:my_label} for comparison in the classification task of images on MNIST, Fashion-MNIST, CIFAR10 and CIFAR100 to explore the capability of the layers' contribution to the performance of the model.

\begin{table}[htbp]
\caption{Models implemented in experiments}
    \centering
    
    \begin{tabular}{|c|c|}
    
    \hline
       Model 1  &  Model 2\\
       \hline
       Convolutional layer, kernel size 4& Convolutional layer, kernel size 4\\
       \hline
      Max pooling (2,2) & Max pooling (2,2) \\
      \hline
      Convolutional layer, kernel size 3 & None\\
      \hline
      Max pooling (2,2) & None\\
      \hline
      Fully connected (128) & Fully connected (128)\\
      \hline
      Fully connected (n classes) & Fully connected (n classes)\\
      \hline
    \end{tabular}
    
    \label{tab:my_label}
    
\end{table}

In the experiments, we have experimented with the generated convolutional layer and the ordinary convolutional layer by replacing the first and second convolutional layers in Model 1 and Model 2 according to the experiment scenario to explore the effectiveness of the generated layers.
\subsection{Supervised Learning of model}

For performance evaluation, we compared the layer generated using approximately 1,500 images to a layer that's fully optimized with the entirety of the available data within a model. In this section, Model 1 serves as the baseline for our assessment.

To explore the effectiveness of the generated layer compared to its counterparts, we have experimented with different freeze settings of the generated layer and compared it to an ordinary layer with the Model 2 trained supervised on all the datasets in this work, as demonstrated below in Fig.~\ref{EXP Supervised ALL} and Fig.~\ref{Metrics_CIFAR10}.
\begin{figure}[htbp]
\centerline{\includegraphics[scale=0.2]{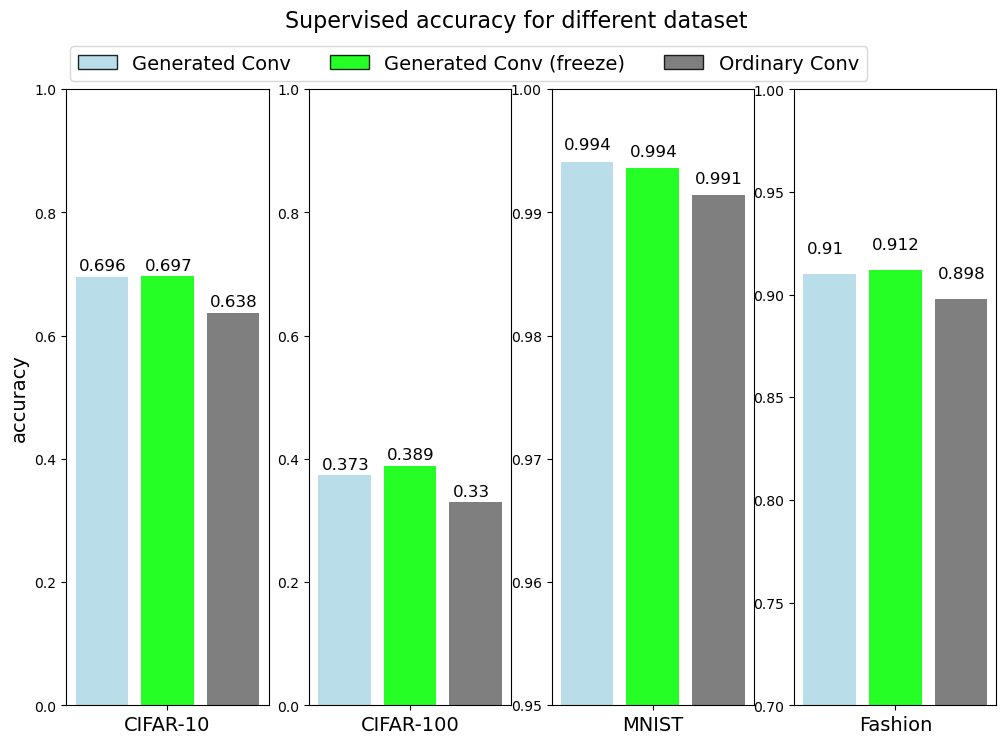}}
\caption{Supervised learning results on 4 different datasets. Results are the highest values for each model taken from 48 random experiments.}
\label{EXP Supervised ALL}
\end{figure}

\begin{figure}[htbp]
\centerline{\includegraphics[scale=0.22]{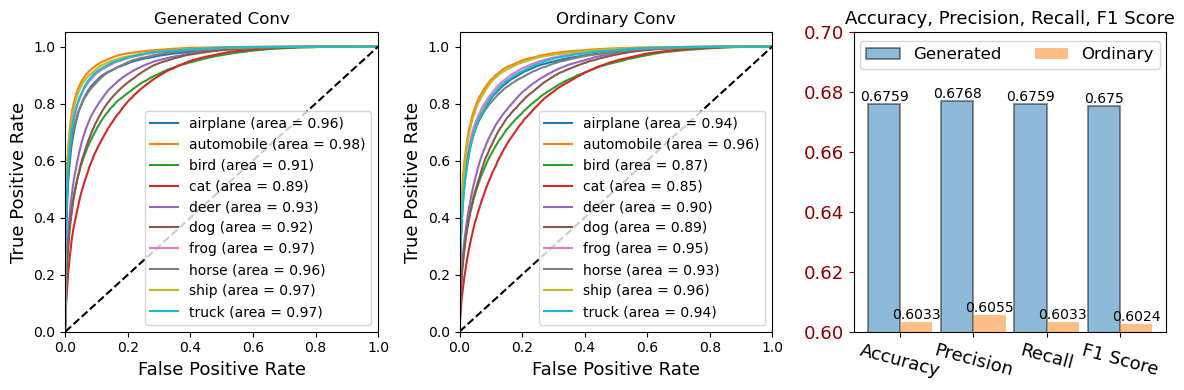}}
\caption{Statistics for 13 randomly initialised models generated or trained on CIFAR10. Generated layers frozen. ROC curves are averaged across 13 models for each class, and accuracy, precision, recall, and f1 score are averaged over 13 models and all classes.}
\label{Metrics_CIFAR10}
\end{figure}

\begin{figure}[htbp]
\centerline{\includegraphics[scale=0.45]{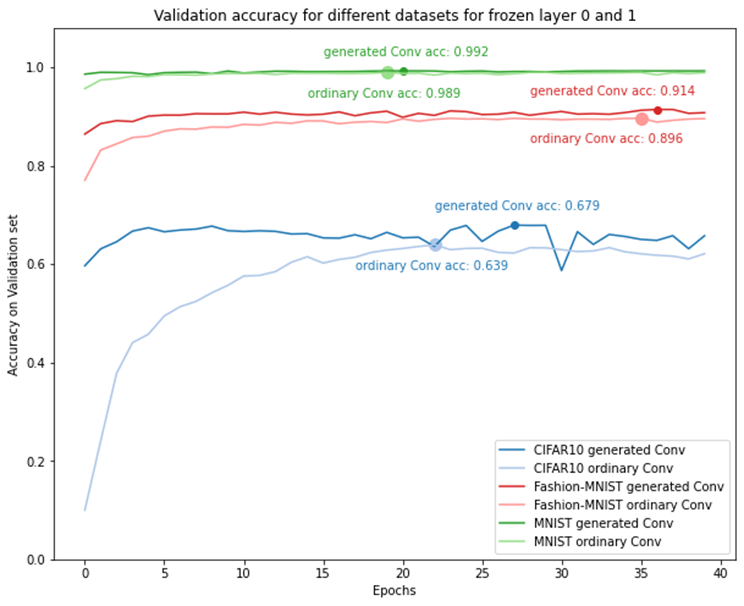}}
\caption{Accuracy of baseline model by substituting both convolutional layers with the generated layer. }
\label{EXP Two layers}
\end{figure}

\iffalse
\begin{table}[htbp]
\caption{Comparison of frozen generated layers to ordinary layer for 21 models on CIFAR10 and CIFAR100}
    \centering
    
    \begin{tabular}{|c|c|c|c|c|}
    
    \hline
         &Accuracy&Precision&Recall&F1 Score\\
       \hline
       Generated Conv&Accuracy&Precision&Recall&F1 Score\\
       \hline
        Ordinary Conv&Accuracy&Precision&Recall&F1 Score\\
        \hline
    \end{tabular}
    
    \label{tab:supervised_CIFAR10}
    
\end{table}
\fi
To show that the generated layer does not conditionally outperform the ordinary convolutional layer by implementing it as the first convolutional layer (layer 0), we have also evaluated its performance by substituting the first and second convolutional layers in the baseline model, with the layer 0 generated first and then layer 1 generated based on the layer 0. The result shows that the generated layer outperforms the ordinary convolutional layer, as shown in Fig.~\ref{EXP Two layers}. These results, shown consistently across 4 different datasets, prove the effectiveness of the generated layer compared to its counterparts.

\subsection{Transfer Learning}
To explore whether increasing the extensible generated layer's size could improve the convolutional layer's performance, we have designed a simple transfer learning experiment by freezing the first layer. \cite{b_transfer_1} By transferring the first layer after training on CIFAR10 to CIFAR100, the results show that the increase in the layer size can bring better overall performance for the frozen layer transferred to a new task.
\begin{figure}[htbp]
\centerline{\includegraphics[scale=0.25]{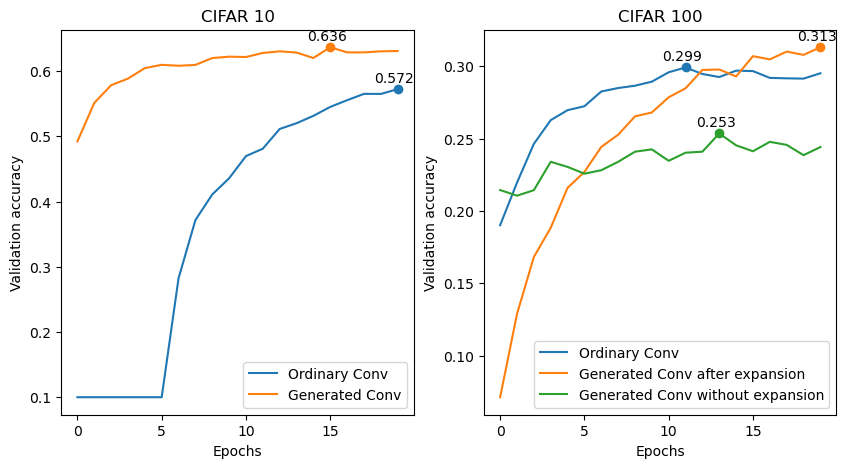}}
\caption{Improve of performance after transferred from CIFAR10 to CIFAR100. Kernels increased from 46 to 84. A lower accuracy of the generated convolutional layer without expansion is due to rejection of patterns from CIFAR100.}
\label{EXP Transfer}
\end{figure}

\section{Discussion and Conclusion}
Our algorithm's major advantage is enabling the expansion of a layer's size within a DNN model based on incoming new data. This feature has brought multiple benefits to the proposed approach, including a more adaptive model (Fig.\ref{EXP Transfer}), fewer data requirements (1,500 images are enough for generation), and compatibility with existing DNNs.

In our experiments, layers we generated consistently enhanced performance when frozen and used in classification models. This is due to each kernel activating its own $P_S$ and intentionally not activating in other patches, as guided by the loss function in section E, resulting in increased sparsity and improved outcomes.

This novel algorithm addresses a significant limitation in the field of deep learning: the static nature of DNN architectures. By dynamising convolutional layers to grow in response to incoming data proactively, we have introduced a more adaptable and efficient approach to constructing a deep learning model. Our experiments have demonstrated the superiority of layers generated by our algorithm over layers trained supervisely. Furthermore, the ability to expand a layer's size based on real-time data enhances the model's adaptability, making it promising for real-world environments with dynamic data streams. As deep learning evolves, methods that offer flexibility, like the one proposed in this study, will be crucial in building models that can seamlessly adapt to ever-changing environments. Future work may delve deeper into refining the algorithm, exploring its applicability in other neural structures, and further optimizing its performance in diverse scenarios.

%\section{Conclusion}

%\section*{References}


\clearpage

\begin{thebibliography}{00}
\bibitem{b1}O’Shea, Keiron and Ryan Nash. “An Introduction to Convolutional Neural Networks.” ArXiv abs/1511.08458 (2015): n. pag.

\bibitem{b1_2}Ren, Shaoqing et al. “Faster R-CNN: Towards Real-Time Object Detection with Region Proposal Networks.” IEEE Transactions on Pattern Analysis and Machine Intelligence 39 (2015): 1137-1149.
\bibitem{b1_3}Cai, Zhaowei and Nuno Vasconcelos. “Cascade R-CNN: Delving Into High Quality Object Detection.” 2018 IEEE/CVF Conference on Computer Vision and Pattern Recognition (2017): 6154-6162.
\bibitem{b1_4}Ronneberger, O., Fischer, P., Brox, T. (2015). U-Net: Convolutional Networks for Biomedical Image Segmentation. In: Navab, N., Hornegger, J., Wells, W., Frangi, A. (eds) Medical Image Computing and Computer-Assisted Intervention – MICCAI 2015. MICCAI 2015. Lecture Notes in Computer Science(), vol 9351. Springer, Cham. https://doi.org/10.1007/978-3-319-24574-4\_28
\bibitem{b1_5}Badrinarayanan, Vijay et al. “SegNet: A Deep Convolutional Encoder-Decoder Architecture for Image Segmentation.” IEEE Transactions on Pattern Analysis and Machine Intelligence 39 (2015): 2481-2495.

\bibitem{b2} Nazir, Maria et al. “Role of deep learning in brain tumor detection and classification (2015 to 2020): A review.” Computerized medical imaging and graphics : the official journal of the Computerized Medical Imaging Society 91 (2021): 101940 .
\bibitem{b2_11}Wu, Wenqi et al. “Face Detection With Different Scales Based on Faster R-CNN.” IEEE Transactions on Cybernetics 49 (2019): 4017-4028.
\bibitem{b2_12}Mao, Longbiao et al. “Deep Multi-Task Multi-Label CNN for Effective Facial Attribute Classification.” IEEE Transactions on Affective Computing 13 (2020): 818-828.
\bibitem{b2_13}Alrashedy, Halima Hamid N. et al. “BrainGAN: Brain MRI Image Generation and Classification Framework Using GAN Architectures and CNN Models.” Sensors (Basel, Switzerland) 22 (2022): n. pag.
\bibitem{b2_14}Aljuaid, Abeer and Mohd Anwar. “Survey of Supervised Learning for Medical Image Processing.” Sn Computer Science 3 (2022): n. pag.
\bibitem{b2_15}Song, Lei et al. “TD-Net:unsupervised medical image registration network based on Transformer and CNN.” Applied Intelligence 52 (2022): 18201 - 18209.

\bibitem{b3} Diehl, Chris and John B. Hampshire. “Real-time object classification and novelty detection for collaborative video surveillance.” Proceedings of the 2002 International Joint Conference on Neural Networks. IJCNN'02 (Cat. No.02CH37290) 3 (2002): 2620-2625 vol.3.

\bibitem{b4} Voulodimos, Athanasios et al. “Deep Learning for Computer Vision: A Brief Review.” Computational Intelligence and Neuroscience 2018 (2018): n. pag.

\bibitem{b2_1} Agarwal, Nidhi et al. “Transfer Learning: Survey and Classification.” (2020).

\bibitem{b2_2} Pan, Sinno Jialin and Qiang Yang. “A Survey on Transfer Learning.” IEEE Transactions on Knowledge and Data Engineering 22 (2010): 1345-1359.

\bibitem{b2_3} De Lange, Matthias et al. “A Continual Learning Survey: Defying Forgetting in Classification Tasks.” IEEE Transactions on Pattern Analysis and Machine Intelligence 44 (2019): 3366-3385.

\bibitem{b2_4} Hospedales, Timothy M. et al. “Meta-Learning in Neural Networks: A Survey.” IEEE Transactions on Pattern Analysis and Machine Intelligence 44 (2020): 5149-5169.

\bibitem{b2_5} Gao, Yuqing and Khalid M. Mosalam. “Deep Transfer Learning for Image‐Based Structural Damage Recognition.” Computer‐Aided Civil and Infrastructure Engineering 33 (2018): n. pag.

\bibitem{b2_6} Mishra, Nikhil et al. “Meta-Learning with Temporal Convolutions.” ArXiv abs/1707.03141 (2017): n. pag.



\bibitem{limitation1}Sattler, Torsten et al. “Understanding the Limitations of CNN-Based Absolute Camera Pose Regression.” 2019 IEEE/CVF Conference on Computer Vision and Pattern Recognition (CVPR) (2019): 3297-3307.
\bibitem{limitation2}Kandi, Haribabu et al. “Incorporating rotational invariance in convolutional neural network architecture.” Pattern Analysis and Applications (2019): 1-14.
\bibitem{limitation3}Yishu Liu et al. "Scene Classification Using Hierarchical Wasserstein CNN." IEEE Transactions on Geoscience and Remote Sensing, 57 (2019): 2494-2509. https://doi.org/10.1109/TGRS.2018.2873966.
\bibitem{limitation4}Gabriel Zaid et al. "Methodology for Efficient CNN Architectures in Profiling Attacks." IACR Trans. Cryptogr. Hardw. Embed. Syst., 2020 (2019): 1-36. https://doi.org/10.13154/tches.v2020.i1.1-36.
\bibitem{limitation5}Chen, Yunhao, et al. "Effective audio classification network based on paired inverse pyramid structure and dense MLP Block." International Conference on Intelligent Computing. Singapore: Springer Nature Singapore, 2023.

\bibitem{b2_n_1} Tammina, Srikanth. “Transfer learning using VGG-16 with Deep Convolutional Neural Network for Classifying Images.” International Journal of Scientific and Research Publications (IJSRP) (2019): n. pag.

\bibitem{b2_n_2} Belouadah, Eden et al. “A Comprehensive Study of Class Incremental Learning Algorithms for Visual Tasks.” Neural networks : the official journal of the International Neural Network Society 135 (2020): 38-54 .

\bibitem{b3_1} Aljundi, Rahaf, Punarjay Chakravarty, and Tinne Tuytelaars. "Expert gate: Lifelong learning with a network of experts." Proceedings of the IEEE conference on computer vision and pattern recognition. 2017.

\bibitem{b3_2} Carpenter, Gail A., and Stephen Grossberg. "Adaptive resonance theory." (2010): 22-35.

\bibitem{b3_2_2} Masuyama, Naoki et al. “Multi-Label Classification via Adaptive Resonance Theory-Based Clustering.” IEEE Transactions on Pattern Analysis and Machine Intelligence 45 (2021): 8696-8712.

\bibitem{b3_3} Terekhov, Alexander V., Guglielmo Montone, and J. Kevin O’Regan. "Knowledge transfer in deep block-modular neural networks." Biomimetic and Biohybrid Systems: 4th International Conference, Living Machines 2015, Barcelona, Spain, July 28-31, 2015, Proceedings 4. Springer International Publishing, 2015.

\bibitem{b3_4} Rusu, Andrei A. et al. “Progressive Neural Networks.” ArXiv abs/1606.04671 (2016): n. pag.

\bibitem{dataset1} Deng, Li. "The mnist database of handwritten digit images for machine learning research [best of the web]." IEEE signal processing magazine 29.6 (2012): 141-142.

\bibitem{dataset2} Xiao, Han, Kashif Rasul, and Roland Vollgraf. "Fashion-mnist: a novel image dataset for benchmarking machine learning algorithms." arXiv preprint arXiv:1708.07747 (2017).

\bibitem{dataset3} Krizhevsky, Alex, and Geoffrey Hinton. "Learning multiple layers of features from tiny images." (2009): 7.

\bibitem{dataset4} Krizhevsky, Alex, Ilya Sutskever, and Geoffrey E. Hinton. "Imagenet classification with deep convolutional neural networks." Advances in neural information processing systems 25 (2012).

\bibitem{b_transfer_1} Kaur, Taranjit and Tapan Kumar Gandhi. “Deep convolutional neural networks with transfer learning for automated brain image classification.” Machine Vision and Applications 31 (2020): n. pag.

\end{thebibliography}
\end{document}